\pdfoutput=1
\documentclass[11pt]{article}

\usepackage[]{naacl2021}

\usepackage{times}
\usepackage{latexsym}
\usepackage{amsmath}

\usepackage{comment}

\usepackage[T1]{fontenc}
\usepackage[utf8]{inputenc}

\usepackage{microtype}

\usepackage{graphicx}

\usepackage{booktabs}
\usepackage{colortbl}
\usepackage{multirow}
\usepackage[skip=2pt]{caption}
\usepackage{textcomp}
\title{Answering Product-Questions by Utilizing Questions from Other Contextually Similar Products}

\author{Ohad Rozen$^1$\thanks{\ \ Work carried out during an internship at Amazon.}, David Carmel$^2$, Avihai Mejer$^2$, Vitaly Mirkis$^2$, and Yftah Ziser$^3$\thanks{\ \ Work carried out while working at Amazon.} 
       \\
       $^1$Computer Science Department, Bar-Ilan University, Ramat-Gan, Israel\\
       $^2$Amazon\\
       $^3$Facebook\\
       {\tt\small ohadrozen@gmail.com, \tt\small \{dacarmel, amejer, vitamin\}@amazon.com, yftahz@fb.com} \\
       }

\iftrue
\newcommand{\avihai}[1]{\noindent{\textcolor{red}{\{{\bf AM:} \em #1\}}}}
\newcommand{\ohad}[1]{\noindent{\textcolor{blue}{\{{\bf OR:} \em #1\}}}}
\newcommand{\david}[1]{\noindent{\textcolor{purple}{\{{\bf DC:} \em #1\}}}}
\newcommand{\yftah}[1]{\noindent{\textcolor{red}{\{{\bf YZ:} \em #1\}}}}
\newcommand{\vitaly}[1]{\noindent{\textcolor{blue}{\{{\bf VM:} \em #1\}}}}
\newcommand{\thomas}[1]{\noindent{\textcolor{green}{\{{\bf TH:} \em #1\}}}}
\fi 

\newcommand{\cpslong}{Contextual Product Similarity}
\newcommand{\cps}{CPS}
\newcommand{\qqlong}{Question-to-Question}
\newcommand{\qq}{Q2Q}
\newcommand{\pqalong}{Product-related Question Answering}
\newcommand{\pqa}{PQA}

\newcommand{\qsolong}{Question Similarity Only}
\newcommand{\qso}{QSO}
\newcommand{\psolong}{Product Similarity Only}
\newcommand{\pso}{PSO}

\newcommand{\paqalong}{Similarity-Based Answer-prediction}
\newcommand{\paqa}{SimBA}
\newcommand{\apc}{APC}
\newcommand{\hybrid}{\paqa+\apc}
\newcommand{\apclong}{Answer Prediction Classifier}
\newcommand{\transformer}{Encoder}

\newcommand{\amazonpqa}{Amazon-PQA}
\newcommand{\amazonpqsim}{Amazon-PQSim}

\newcommand{\capsize}[1]{\footnotesize #1}

\newcommand{\lightbars}{Light bars}
\newcommand{\monitors}{Monitors}
\newcommand{\smartwatches}{Smartwatches}
\newcommand{\receivers}{Receivers}
\newcommand{\backpacks}{Backpacks}
\newcommand{\jeans}{Jeans}
\newcommand{\beds}{Beds}
\newcommand{\home}{Home \& office desks}
\newcommand{\masks}{Masks}
\newcommand{\posters}{Posters \& prints}
\newcommand{\accessories}{Accessories}

\begin{document}
\maketitle
%\david{\bf{Alternative title:} Answering a Product Question by Similar Resolved Product Questions}

\begin{abstract}
Predicting the answer to a product-related question is an emerging field of research that recently attracted a lot of attention. Answering subjective and opinion-based questions is most challenging due to the dependency on customer-generated content. Previous works mostly focused on review-aware answer prediction; however, these approaches fail for new or unpopular products, having no (or only a few) reviews at hand. 
In this work, we propose a novel and complementary approach for predicting the answer for such questions, based on the answers for similar questions asked on similar products. We measure the contextual similarity between products based on the answers they provide for the same question. A mixture-of-expert framework is used to predict the answer by aggregating the answers from contextually similar products. Empirical results demonstrate that our model outperforms strong baselines on some segments of questions, namely those that have roughly ten or more similar resolved questions in the corpus. We additionally publish two large-scale datasets\footnote{The datasets are freely available at \url{https://registry.opendata.aws} under the names \textit{Amazon-PQSim} and \textit{Amazon-PQA}.} used in this work, one is of similar product question pairs, and the second is of product question-answer pairs. 
\end{abstract}

\section{Introduction}
\label{sec:intro}
\pqalong{} (\pqa) 
%\david{PQA? I suggest to take out the Community, PQA is the common term.}
is a popular and essential service provided by many e-commerce websites, letting consumers ask product related questions to be answered by other consumers based on their experience. 
The large archive of accumulated resolved questions can be further utilized by customers to support their purchase journey and automatic product question answering tools (e.g. \newcite{jeon2005finding,cui2017superagent,Carmel2018}).
However, there are many unanswered questions on these websites, either because a newly issued question has not attracted the community attention yet, or because of many other reasons ~\cite{DBLP:conf/sigir/ParkKZG15}. This may frustrate e-commerce users, in particular when their purchase decision depends on the question's answer. Automatic %\pqalong{} (\pqa) 
\pqa{} may assist the customers and the sellers by answering these unanswered questions, based on various diversified resources.

Previous \pqa{} approaches leverage product specifications and description information~\cite{cui2017superagent,Lai2018,DBLP:conf/wsdm/GaoRZZYY19}, as well as customer-reviews~\cite{yu2012answering,DBLP:conf/www/McAuleyY16,DBLP:conf/wsdm/YuL18,das2019learning,fan2019reading,chen2019answer,Deng2020}, for answering product related questions. However, there are two notable shortcomings to these two approaches. Product information can typically address questions about product features and functionality, but can't address complex and subjective questions such as opinion question ({\em Is it good for a 10 year old?}), advice-seeking question ({\em What is the color that best fit my pink dress?}), or unique usage questions ({\em Can I play Fifa 2018 on this laptop?}). Customer-reviews, on the other hand, can partially address this kind of questions~\cite{wan2016modeling}, yet there are many products with few or no reviews available, either because they are new on the site or are less popular.

We propose a novel and complementary approach for answering product-related questions based on a large corpus of \pqa{}.
Given an unanswered product question, we seek similar resolved questions\footnote{We consider questions \textit{similar} if they have the same semantic intent. For example, {\em can I wash this?}, {\em Is the product washable?}, {\em Is it ok to clean it with water?} are all considered as similar questions when asked in context of a similar product.} about similar products and leverage their existing answers to predict the answer for the customer's question. We call our method \paqa{} (\textbf{Sim}ilarity \textbf{B}ased \textbf{A}nswer Prediction).
For example, the answer for the question {\em ``Will these jeans shrink after a wash?''},
asked about a new pair of jeans on the website, 
may be predicted based on the answers for similar questions asked about other jeans that share properties such as fabric material, brand, or style. An example is shown in Table~\ref{tab:Sim_q_cx}.
The main hypothesis we explore in this work is whether the answer to a product question can be predicted, based on the answers for similar questions about similar products, and how reliable this prediction is.
% Table \ref{tab:Sim_q_cx} demonstrates how an answer to a question about a jeans pair can be predicted based on the answers for the same question asked about other similar jeans in the market.
% Figure \ref{fig:Sim_q_cx} demonstrates how an answer to a question about a smart-watch can be predicted based on the answers for the same question asked about many other similar smart-watches in the market.\ohad{not accurate. the figure doesn't explain \textbf{how} it can be predicted. It just demonstrates the use case}

As our method relies on the existing \pqa{} corpus, it addresses the two mentioned shortcomings of the previous approaches. First, it can address a variety of product-related questions that are common in \pqa{}, including subjective and usage questions. Second, our method can provide answers to new or less popular products as it leverages an existing set of similar questions from other similar products.

A key element of our proposed method is a novel concept that we refer to as \cpslong{}, which determines whether two products are similar in the context of a specific question. For example, two smart-watches may be similar with regards to their texting capability but different with regards to sleep monitoring. 
%cellular connectivity.
%\david{Any Better example?} \avihai{is this better?}
In Section~\ref{sec:methodology} we formally define this concept and propose a prediction model for measuring contextual similarity between products, with respect to a given question. Additionally, we describe an efficient method to train this model by leveraging an existing \pqa{} corpus.

Another appealing property of \paqa{} is its ability to support the predicted answer by providing the list of highly similar questions upon which the answer was predicted, hence increasing users' confidence and enhancing user engagement. 
% An illustration of a potential user interface is depicted in Figure~\ref{fig:Sim_q_cx}.

Our main contributions are: (a) A novel \pqa{} method that overcomes several shortcomings of previous methods. (b) A novel concept of \cpslong{} and an effective way to automatically collect annotations to train this model. 
% (c) \avihai{We should merge this contribution with (a) - a novel method along with empirical evaluation.} 
% (c) Empirical evaluation of our method which demonstrates that it outperforms a strong baseline in some question segments, and that a hybrid model is effective. 
(c) Finally, publishing two large scale datasets, one is a question similarity data set and the second is
%%%TODO: \ohad{the world's largest}%%%
a large-scale Amazon product questions and answers dataset, details are provided in Section~\ref{sec:datasets}. 

Empirical evaluation of our method demonstrates that it outperforms a strong baseline in some question segments, and that a hybrid model is effective in all the vast majority of the questions. 

% \begin{figure}
%   \centering
%   \includegraphics[width=3in]{Figs/Similar_Q_CX.png}
%   \caption{\capsize{Illustration of a potential user interface, supporting the predicted answer by the list of highly similar questions, together with their answers.}\ohad{I think displaying only YES answers is more representative of our use case (we'd like to show \textbf{supporting} answers)}}
%   \label{fig:Sim_q_cx}
% \end{figure}

\begin{table}[!t]
    \centering
    \footnotesize
    \begin{tabular}{l}
    \toprule
    \textbf{Product}: Dickies Men's Jeans, 100\% Cotton. \\
    ~~~\textbf{Q:} Will these shrink after a wash? \\ \ \ \ \textbf{Predicted answer:} No \\ 
    \hline
    \textbf{Similar Product 1:} Eddie Bauer Men's Jeans, 100\% \\Cotton. \\
    ~~~\textbf{Q:} Do these shrink when you wash and dry them?  \\
    ~~~\textbf{A:} No \\
    \textbf{Similar Product 2:} Timberland PRO Men's Jean, 99\% \\Cotton, 1\% Polyurethane. \\
    ~~~\textbf{Q:} Was there any shrinkage after washing? \\
    ~~~\textbf{A:} No shrinkage \\
    \textbf{Similar Product 3:} Levi's Men's Jeans, 98\% Cotton, \\2\% Elastane. \\
    ~~~\textbf{Q:} Do these shrink at all during washing/drying? \\
    ~~~\textbf{A:} They have not shrunk \\
    \bottomrule
    \end{tabular}
    \caption{\capsize{Answer prediction example based on similar questions asked about similar products. The answer for all \textit{contextually-similar} products is `\textit{no}' therefore we predict the answer `\textit{no}' for the target question.}}
    \label{tab:Sim_q_cx}
\end{table}

\section{Related Work}
\label{sec:related_work}
Automatic aswering product related questions has become a permanent service provided by many e-commerce websites and services \citep{cui2017superagent, Carmel2018}. Questions are typically answered based on product details from the catalog, existing Q\&A's on the site, and  customer reviews. Each of these resources, used for answer generation, has been studied extensively by the research community recently, probably due to the complexity of this task, the availability of appropriate datasets \citep{amazon-qa-data}, and the emergent increase in on-line shopping usage.

\newcite{Lai2018} built a question answering system based on product facts and specifications. They trained a question answering system by transfer learning from a large-scale Amazon dataset to the Home Depot domain.
\newcite{DBLP:conf/wsdm/GaoRZZYY19} generated an answer from product attributes and reviews using adversarial learning model which is composed of three components: a question-aware review representation module, a key-value attribute graph, and a seq2seq model for answer generation.
\newcite{yu2012answering} answered opinion questions by exploiting hierarchical organization of consumer reviews, where reviews were organized according to the product aspects.

The publication of Amazon datasets of reviews\footnote{https://nijianmo.github.io/amazon/index.html}
 and Q\&As~\cite{amazon-qa-data}, triggered a flood of studies on review-aware answer prediction and generation. 
\newcite{DBLP:conf/www/McAuleyY16} formulated the review based question answering task as a mixture-of-experts framework --- each review is an ``expert'' that votes on the answer to a yes/no question. Their model learns to identify `relevant' reviews based on those that vote correctly. In a following work, \newcite{wan2016modeling} observed that questions have multiple, often divergent, answers, and the full spectrum of answers should be further utilized to train the answering system.

\newcite{chen2019answer} described a multi-task attention mechanism which exploits large amounts of Q\&As, and a few manually labeled reviews, for answer prediction.
\newcite{fan2019reading} proposed a neural architecture, directly fed by the raw text of the question and reviews, to mark review segment as the final answer, in a reading comprehension fashion.
\newcite{das2019learning} learned an adversarial network for inferring reviews which best answer a question, or augment a given answer.
\newcite{Deng2020} incorporated opinion mining into the review-based answer generation.
\newcite{DBLP:conf/wsdm/YuL18} generated aspect-specific representation for questions and reviews for answer prediction for yes-no questions.
\newcite{Yu2018} used transfer learning from a resource-rich source domain to a resource-poor target domain, by simultaneously learning shared representations of questions and reviews in a unified framework of both domains.

%%%%%%%%%%%%%%%%%%%%%%%%%%%%%%%%%%%%%%%%%%%%%%%
%\textbf{Question answering.}
%What is the SOTA for question answering? Can we claim the RoBERTa model we used is a strong enough baseline?

%\textbf{\pqa} Previous works that address \pqa~ utilize the product information. The \emph{\textbf{Mdqa}} method in \cite{DBLP:conf/www/McAuleyY16} utilize the free text product description, while \cite{DBLP:conf/wsdm/GaoRZZYY19}, \cite{FWENLPTuan18} rely on product specifications in a structured attribute-value format. One limitation of product information and specifications is that it does not address subjective or unique customer questions such as "Can I play Fifa 2018 on this laptop?". Our method, which relies on \cqa~ can address this kind of questions.

%Other methods utilize customer reviews \cite{DBLP:conf/www/McAuleyY16}, \cite{DBLP:conf/wsdm/YuL18}, \cite{DBLP:conf/sdm/FanFSLW19} and can address subjective and unique questions. However, one notable limitation is that new and less popular products have only few or even no available customer reviews. Our approach does not rely on customer reviews, instead, we utilize \cqa~ of other products, and therefore can provide answers even for new products.

All this line of works assume the existence of rich set of product reviews to be used for question answering. This solution fails when no reviews are available. The challenge of review generation for a given product, while utilizing similar products' reviews, was addressed by \newcite{DBLP:conf/sigir/ParkKZG15}. For a given product they extracted useful sentences from the reviews of other similar products.
Similarly, \cite{Pourgholamali16} mined relevant content for a product from various content resources available for similar products. Both works focused on the extraction of general useful product related information rather than answering a specific product  question, as in our case.  Second, the product-similarity methods they considered rely on product specifications and description, and do not depend on the question to be answered, while our method considers a specific question at hand when estimating contextual product similarity.

\section{Similarity-Based Answer Prediction}
\label{sec:methodology}
\begin{figure}
   \centering
   \includegraphics[width=3in]{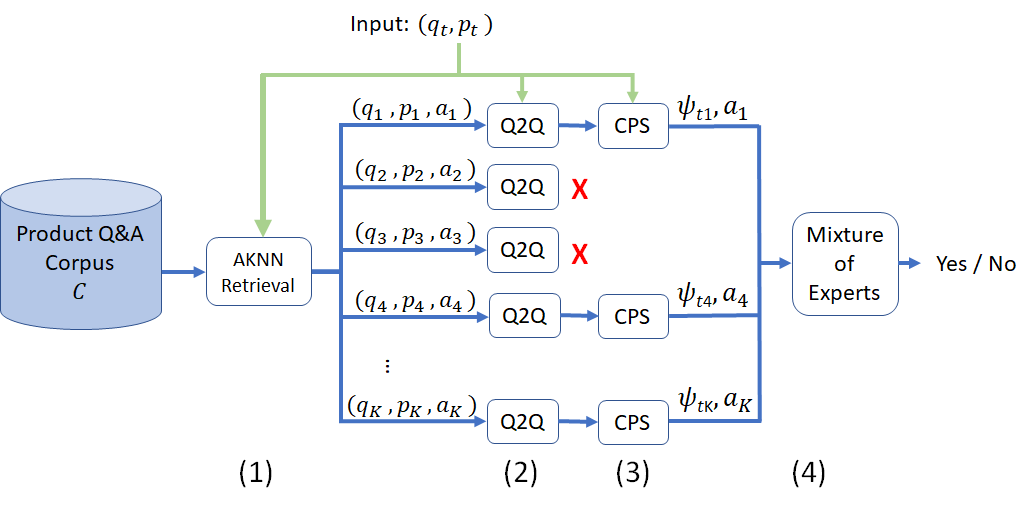}
   \caption{\capsize{Overview of \paqa{} answer prediction framework. (1) K siblings to the product question are retrieved from corpus by AKNN. (2) Siblings are filtered by the Q2Q model keeping only twins. (3) Twins are scored by the CPS model. (4) A Mixture of Experts model uses these votes to predict the answer.}}
   \label{fig:method}
\end{figure}

In this section, we introduce the \paqalong{} (\paqa) method for predicting the answer for a product question, based on the answers for other similar product questions.
%\david{too early: yes-no, product questions in a cold-start setting}. 
We restrict our study to yes/no questions only, due to their popularity in the \pqa{} domain (54\% on our PQA dataset), and following common practices in answer prediction studies  \cite{DBLP:conf/www/McAuleyY16,DBLP:conf/wsdm/YuL18}.
%Our main contribution is utilizing the existing products and their corresponding Q\&As as a knowledge-base rather than merely a training set, thus using answers to similar questions asked on similar products as our signal. 
Figure~\ref{fig:method} presents our prediction framework and its main components.
%, described in details in the following.

%DC: Formally, we denote the record of a target (unanswered) question $q_t$, asked about a product $p_t$, as 2-tuple record $\rect=(q_t, p_t)$. The existing question-product-answer 3-tuple records are signified by $\recc_j= (q_j,p_j,a_j) \in C$, where $q_j$, $p_j$, and $a_j$ are the question, product and answer or the record, $C$ signifies the corpus of the existing records, and $j$ is the index with $1 \leq j \leq |C|$. We treat $C$ as a knowledge-base to be used for our answer prediction task. 
Formally, a question-product-answer tuple is denoted by $r_j=(q_j, p_j ,a_j)$, where $a_j\in\{'yes','no'\}$. $C=\{r_j\}_{j=1}^N$ is the set of $N$ tuples of a given product category. 
$r_t=(q_t,p_t, ?)$\footnote{The answer for the target record is unknown.} is the target record of an unanswered question $q_t$, asked about product $p_t$. We treat $C$ as the knowledge-base we use for answering $q_t$.

%DC In order to predict the yes-no answer $a_t$ for $q_t$, we first retrieve records $\{\recc_{k_1}, \recc_{k_2}...\}$ from $C$, with questions $\{q_{k_1}, q_{k_2}...\}$ that are similar to $q_t$ (Section~\ref{sec:knn_method}).
Given a target record $r_t$, in order to predict its answer $a_t$, we first retrieve a set of records from $C$ with the most similar questions to $q_t$ (Figure~\ref{fig:method}, stage 1).
We denote the retrieved records as \textit{siblings} of $r_t$. 
We then filter the siblings by applying a Question-to-Question similarity (\qq) model, keeping only records with highly similar questions which are expected to have the same question intent as of $q_t$, (Figure~\ref{fig:method}, stage 2). We denote these records as \textit{twins} of $r_t$. 
We then apply our Contextual Product Similarity (\cps{}) model to measure the \textit{contextual} similarity between $r_t$ and its twins (Figure~\ref{fig:method}, stage 3). 
The \cps{} similarity score is used to weight the twins by considering them as voters, applying a mixture-of-experts model over their answers for the final answer prediction (Figure~\ref{fig:method}, stage 4). More details about the model's components, the training processes, and other specifications, are described in the following. 
%and in Section~\ref{sec:experimental_setup}.

\subsection{Sibling Retrieval}
\label{sec:knn_method}
%DC Given a corpus $C=\{\recc_1 .. \recc_N\}$ of all records of resolved yes/no questions in a product category, our first goal is to retrieve all records having a question with the same intent as of $q_t$.
Given a target record $r_t$, and a corpus of product-question-answer records $C$, our first goal is to retrieve all records with a question having the same intent as of $q_t$.
%Since $C$ contains many question-product-answer records, 
As $C$ might be very large, 
applying a complex neural model to measure the similarity of each question in $C$ to $q_t$ is often infeasible. We therefore apply a two step retrieval process. 
In a preliminary offline step, we index the records in $C$ by creating embedding vectors  for their questions,
 %$\{q_j \in r_j^c}$ 
 using a pre-trained encoder. For retrieval, done both during training and inference, we similarly embed the question $q_t$ into vector $e_t$. 
% should we say USE explicitly? or should we leave it to the Experiments section/we use a fast KNN method to retrieve the top K most similar questions together with their related products. For that, 
We then use a fast Approximate K Nearest Neighbors (AKNN) search to retrieve $K$ records, with the most similar questions,
%, with their related records $S_{\rect} = \{\recc_{k_1}..\recc_{k_K}\}$, 
based on the cosine similarity between $e_t$ and the embedding vectors of the questions in $C$. We denote the set of retrieved siblings of $r_t$ by $S(r_t)$.
% \footnote{While we are aware that there are more advanced sentence similarity approaches that can be applied for this task \ohad{[cite{}]}, our implementation is considerably faster.
%  It takes $O(N)$ for creating the embedding vectors, and $O(log(N))$ to retrieve top K results for one target question, which we can better refine with a more time-consuming superior model.\avihai{I think we don't need this comment at all. We can control the speed-vs-accuracy by changing the K in the KNN. Setting K=1M would return ALL the corpus for the Q2Q step.}}.
 
\subsection{Twin detection}
\label{sec:q2q_method}
The retrieved sibling records are those with the most similar questions to the target question.
In the second step of the retrieval process, we enhance our record selection by applying a highly accurate transformer-based \qqlong{} (\qq) classifier (See Section~\ref{sec:data_preparation}), which we train over our question to question similarity dataset (Section~\ref{sec:pqsim}). 
The $Q2Q(q_t,q_k)$ classifier predicts the similarity between a target question $q_t$ and each of the questions $q_k$ in 
%DC $S_{\rect}$, 
$S(r_t)$.
% DC yielding a similarity probability $S_{\qq}(q_t, q_k)$. 
%DC $\recc_k$ is considered a \textit{twin} of $\rect$ if $S_{\qq}(q_t, q_k) >\gamma$, where $0.5\leq\gamma\leq 1.0$ is a hyper-parameter of the system. We denote the set of indices $\{j_{k_0},j_{k_1},...\}$ of all twins of $\rect$ as $T_{\rect}$. 
A record $r_k$ is considered a \textit{twin} of $r_t$ if $\qq(q_t, q_k) >\gamma$, where $0.5\leq\gamma\leq 1.0$ is a hyper-parameter of the system. We denote the set of twins of $r_t$ by $T(r_t)$.

\subsection{Contextual Product Similarity (CPS)}
\label{sec:cps_method}
We consider products $p_1$ and $p_2$ to be contextually similar, with respect to a yes/no question $q$, if the answer to $q$ on both products is the same\footnote{By design, both products belong to the same product category $C$, which prevents comparing unrelated products. For example, comparing an airhorn and a computer fan in the context of the question \textit{is it loud} is therefore prevented.}.
Given a pair of twin records $(r_1, r_2)$,  our \cps{} model is aims to predict the contextual similarity between them, i.e. whether their (highly similar) questions have the same answer.

Since $r_1$ and $r_2$ are twins, 
their questions are expected to have the same intent; yet, they might be phrased differently. To avoid losing any information, we provide both questions as input to the \cps{} model, during training and during inference time.

\begin{figure}
   \centering
   \includegraphics[width=2.2in]{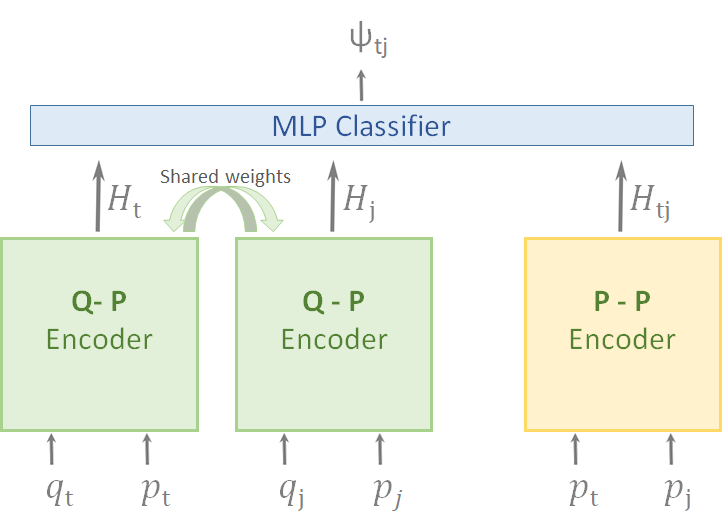}
   \caption{\capsize{The Contextual Product Similarity (CPS) model. The target question-product pair $(q_t, p_t)$ and the twin question-product pair $(q_j, p_j)$ are encoded using a transformer encoder, while the questions attend the product text. The texts of both products are coupled and also encoded, allowing the two product text attend each other. The three output vectors are then concatenated and classified using an MLP classifier.}}
   \label{fig:cps_model}
\end{figure}

\paragraph{\cps{} Model Architecture}
 Figure~\ref{fig:cps_model} depicts the \cps{} model for predicting the contextual similarity
 between a target record $r_t$, and one of its twins - record $r_j$.
For each record, the question-product pair is embedded using a pre-trained transformer encoder, allowing the product textual content and the question text attend each other\footnote{The product textual content can be accumulated from several resources. In our experiments, we restrict the product content to its title and bullet points.}: %\avihai{we are very general here, but in the experiment section need to specify the exact text content we used} 
\begin{align*}
\mathbf{H_t}= \transformer(q_t, p_t), ~~~ 
\mathbf{H_j}= \transformer(q_j, p_j)
\end{align*}
The two models share weights to avoid over-fitting and for more efficient learning. 
A second encoder embeds the textual content of both products, encapsulating the similarity between them:
\begin{align*} 
\mathbf{H_{tj}} &= \transformer(p_t, p_j) 
\end{align*}
Then, a one hidden  MLP layer takes the concatenation of the three embedding vectors, to predict the probability of $a_t=a_j$,
\begin{equation}
%\small
\begin{array}{ll}
\psi_{tj} = \cps(r_t,r_j)=P(a_t\text{=}a_j|r_t, r_j)&\\
 = MLP(\mathbf{H_t}\oplus \mathbf{H_j} \oplus\mathbf{H_{tj}}) &
\end{array}
\label{eq:4}
\end{equation}

%\paragraph{CPS Benefits???} \ohad{Any Ideas for a better title ??? 2 paragraphs to be represented by the title: 1 the concept of contextual similarity 2. labeling in scale }

%Our contextual similarity approach allows two products to consider \textit{similar} in the context of one question and \textit{different} in the context of another. In Section~\ref{sec:experiments}, we show how this ability empowers the model and leads to improved prediction performance.\ohad{Anyone has an explanation why it's important and empowers...? Avihai?}

Another key advantage of the \cps{} model is its ability to be trained on a large scale, without human annotations, by simply yielding the training labels directly from the polarity between the answers of twin pairs extracted from our training data. For any pair of twins $(r_i, r_j)$:
\begin{equation}
    \textrm{label} (r_i, r_j)= 
\begin{cases}
    \textrm{similar},& a_i=a_j\\
    \textrm{different},& a_i\neq a_j
\end{cases}
\label{eq:labeling}
\end{equation}

\begin{table*}[!t]
    \centering
    \small
    
    \begin{tabular}{lll}
    \toprule
    \textbf{Question 1}  & \textbf{Question 2}  &  \textbf{Label} \\
    \midrule
    Can this be used with regular light bulbs? & Can i put a regular light bulb in this? & Similar \\
    Can i use these labels in a laser printer? & Can this be used in laser printer? & Similar \\
    Does the hat have an adjustable strap inside? & Does the hat adjust inside? & Similar \\
    Can this organizer hold sleeved cards? & Can it hold cards with sleeves on them? & Similar \\
    Does this phone have bluetooth? & Does the phone have gps? & Different \\
    Can just two player play this game & Whats the length of play of this game? & Different \\
    Is there a diffuser for this dryer? & Can this go in the dryer? & Different \\
    What material is the neck strap made of? & Does this come with a neck strap? & Different \\
    \bottomrule
    \end{tabular}

    \caption{Examples from Amazon-PQSim Dataset. Each example consists of a user-generated question pair and a human-annotated label for their similarity.}   
    
    \label{tab:pqsim_examples}
\end{table*}

\subsection{Mixture of Experts}
\label{sec:mixure_of_experts_method}
 A mixture of experts is a widely-used method to combine the outputs of several classifiers by associating a weighted confidence score with each classifier \cite{mixture_of_experts}. In our setting, experts are individual twins that lend support for or against a particular answer for a question. Each twin is weighted by its contextual similarity to the target record $r_t$, as predicted by the \cps{} model. 
 %Specifically, any twin  $r_j \in T(r_t)$  is weighted by $\psi_{tj}^2$. 
% %Let $r_j$ be a twin of $r_t$, then $r_j$ is weighted by  $\psi_{tj}^2$. 
% Additionally, we apply a lower weight-limit $w_{min}$ for all twins, where $0 \leq w_{min} \leq 0.5 $ is a hyper-parameter that we tune on the development set.\footnote{We tried using the \cps{} score for all twins, i.e. $w_{min}=0$, however, using a fine-tuned minimal weight yielded better results.} 
% We consider all twins, even the non-similar ones, to gain a better prior estimation of the answer.

% To mitigate errors of the \cps{} model, we also consider twins with a lower contextual similarity score $\psi_{tj}$
% \avihai{how about a different motivation instead of mitigating errors: we consider all twins, even the non-similar, to gain a better prior estimation of the answer}
% % \ohad{we need to explain why non-similar twins can help the prediction, which by definition can't. it's only the errors} 
% by applying a lower weight-limit $w_{min}$ for all twins, where $0 \leq w_{min} \leq 0.5 $ is a hyper-parameter that we train on the development set.\footnote{We tried using the \cps{} score for all twins, i.e. $w_{min}=0$, however, using a fine-tuned minimal weight yielded better results.} 

Given a target record $r_t$, the weight of each of its twins, $r_j \in T(r_t)$ is determined by 
\begin{align*} 
\lambda(r_j)= \textrm{max}(\psi^2_{tj}, w_{min})
\end{align*}
where $\psi_{tj}=CPS(r_t,r_j)$, and $0 \leq w_{min} \leq 0.5 $ is a lower weight-limit; a hyper-parameter that we tune on the development set.\footnote{We tried using the \cps{} raw score for all twins, i.e. $w_{min}=0$, however, using a fine-tuned minimal weight yielded better results.}

% The probability of the predicted answer $a_t=\textrm{`yes'}$ is therefore derived by \avihai{change to sign(sigma)}
% \begin{align} 
%     P(a_t|r_t)=\frac{1}{2} + \frac{1}{2|T(r_t)|}
%     \sum_{r_j\in T(r_t)}
%     \lambda(r_j)\delta(a_j)
%     \label{eq:mixture_eq}
% \end{align}
The predicted class of $a_t$ is therefore derived by 
\begin{equation} 
    Pred(a_t|r_t)=\textrm{sign} \left(\sum_{r_j\in T(r_t)} \lambda(r_j)\delta(a_j) \right)
    \label{eq:mixture_eq}
\end{equation}
where positive/negative $Pred$  indicates `yes'/`no' respectively, and
$\small
    \delta(a) = 
    \begin{cases}
    +1, &  a=\textrm{`yes'}\\
    -1,&  a =\textrm{`no'}.
\end{cases}
$

Our methodology can be easily expanded to incorporate more answer predictors (voters) of different types into \paqa{}. An example for such an expansion is described at Section~\ref{sec:baselines}.

\section{Datasets}
\label{sec:datasets}

We introduce two new datasets to experiment with our answer prediction approach: 1) The Amazon Product Question Similarity (Amazon-PQSim) dataset which is used to train our \qq{} model; 2) The Amazon Product Question Answers (Amazon-PQA) dataset of product related Q\&As, used for training the \paqa{} model.

\subsection{Amazon-PQSim Dataset}
\label{sec:pqsim}
We collected a first-of-a-kind question-to-question similarity dataset of product-question pairs from the Amazon website (Amazon-PQSim. See Table~\ref{tab:pqsim_examples} for examples). Unlike the Quora dataset of general question pairs\footnote{https://www.kaggle.com/c/quora-question-pairs}, product questions are asked in the context of a designated product page. This makes them unique and different from questions asked in other domains. For example, the question {\em Is it waterproof?}, when appears on the {\em Fitbit Flyer} detailed page, should implicitly be interpreted as {\em Is Fitbit Flyer waterproof?}. 

The following steps were taken for the data collection: (a)  randomly sampling product-questions from the Amazon website. (b) filtering out some of these questions (e.g., non-English questions, for more details, see Appendix~\ref{sec:appendix_a}).
(c) For each of the remaining questions, we retrieved up to three candidate similar questions from the collection. A question is paired with the original question if the Jaccard similarity among them is in the range of $[0.3, 0.5]$ . We ignore highly similar questions ($ >0.5$) since we don't want nearly verbatim pairs in our dataset, as well as  dissimilar pairs ($< 0.3$). 
(d) Finally we used the Appen crowd-sourcing platform\footnote{https://appen.com} for manual annotation of question pairs similarity \footnote{As the questions are asked in context of a specific product, they are often written in an anaphoric form (e.g. \textit{Is it waterproof?}). To keep our dataset general, we instructed the judges to accept such questions as if they included the actual related product name. For example, the pair \textit{Is it waterproof?} and \textit{Is this Fitbit waterproof?} were labeled as \textit{similar}.}. Each question pair was labeled by at least three judges, and up to seven, until reaching agreement of 70\% or more.

The above steps resulted in a nearly balanced dataset (1.08 positive-negative ratio) of more than 180K product question pairs with judges agreement of 70\% or more, and among them about 90K question pairs have perfect judges agreement (1.14 positive-negative ratio).

\subsection{\amazonpqa{} Dataset}
\label{sec:pqa}
We collected a large corpus of product questions and answers from the Amazon website, similar to 
the popular Amazon Q\&A dataset~\cite{amazon-qa-data}.
%the one described at \citep{DBLP:conf/www/McAuleyY16}. 
%We aim to collect all available questions per narrow sub-category, as oppose to a sample across a broad set of categories, as done in the popular Amazon question/answer data\cite{amazon-qa-data}.
Since our answer prediction method directly utilizes an existing corpus of resolved questions, we aim to collect {\em all} available questions per narrow sub-category instead of a sample of questions across broad categories by the popular Amazon Q\&A dataset. For example, instead of sampling from the broad Electronics category, we collect all questions under the narrower \textit{\monitors{}} and \textit{\receivers{}} categories. 
% While our dataset contains questions from all categories, in our experiments we focus on questions from eleven narrow categories while considering yes-no questions only. 

\paragraph{Raw Data Extraction}
We collected all product questions, with their answers, from 100 sub-categories, available on the Amazon website in August 2020.
Overall, 10M questions were collected, with 20.7M answers, on 1.5M products. For full statistics of the raw data, see Table~\ref{tab:extracted_data_stats} in Appendix~\ref{sec:appendix_a}.

% We collected all product questions, with their answers, from 11 sub-categories, available on the Amazon website in August 2020.
% Overall, 1.1M questions were collected, with 2.1M answers, on 188K products. For full statistics of the raw data, see Appendix~\ref{sec:appendix_a}.
%\footnote{Only products that were alive at the time of the snapshot were included \david{what do you mean alive? seems like TMI.}.}.

\paragraph{Yes/No Question Classification} 
We followed  \cite{google-he} for detecting Yes/No questions using simple heuristics. See Appendix~\ref{sec:appendix_a} for details.

% , such as checking if the question starts with a \textit{Be verb} (am, is, are, been, being, was, were), \textit{Modal verb} (can, could, shall, should, will, would, may, might) or an \textit{Auxiliary verb} (do, did, does, have, had, has), and additionally ends with a question mark. We tested the classifier on  McAuley's dataset~\cite{amazon-qa-data}, 
% identified yes/no questions with 98.4\% precision at 96.5\% recall, while considering McAuley as ground truth\footnote{\citeauthor{DBLP:conf/www/McAuleyY16} reported identifying yes/no questions with 97\% precision at 82\% recall on their dataset.}.%\ohad{Not sure if it's ok to state that we used McAuley as ground truth}
% achieving 97.7\% accuracy while considering this dataset as a ground truth. 

\paragraph{Yes/No Answer Labeling} 
Questions are typically answered by free-text answers, posted independently by multiple users. In order to convert these answers into a single yes/no answer, we first classified each answer into one of three classes: \textit{yes}, \textit{no} and \textit{maybe}, and then used majority vote among the classified answers. We used a pre-trained RoBERTa-based classifier, and trained the model on McAuley's dataset~\cite{amazon-qa-data}, taking only yes/no questions. 
% To train the model, we used the McAuley dataset~\cite{amazon-qa-data}, taking only yes/no questions. 
%We determine each question’s final yes/no answer. If an answer is provided by a verified seller we used it as the final label. Otherwise we used the majority vote among the answers, ignoring \textit{maybe} answers. We also remove questions with equal number of yes and no answers.
See Appendix~\ref{sec:appendix_a} for details.

%%%% Table 2 - Statistics of the Amazon-PQA -Train-Dev-Text - average only
% \begin{table}
%     \centering
%     \small
%     \input{Figs/pqa_stats_av}
%     \caption{\capsize{Statistics of the Amazon-PQA dataset. \textit{\%  Yes} is percentage of records with a \textit{yes} answer. *Macro Average.}}
%     \label{tab:pqa_stats_av}
% \end{table}

\section{Experiments}
\label{sec:experiments}
%\avihai{Provide some overview of the entire experimental section: We now describe the }
We experiment with eleven product categories covered by our Amazon-PQA dataset (Section \ref{sec:pqa}), training a \paqa{} answer prediction model for each of the categories independently. Next, we describe the data preparation steps for each of the \paqa{} components.

\subsection{Data Preparation}
\label{sec:data_preparation}
%DC (???) As described in Section \ref{sec:cps_method}, each input example for the CPS model consists of two twin records, $\recc_i$ and $\recc_j$, while $q_i$ and $q_j$ must be similar questions. We use our PQA training set as the corpus $C$ and build an auxiliary dataset $D_{train}$ using the two retrieval steps described in Sections \ref{sec:knn_method} and \ref{sec:q2q_method}:

\paragraph{Sibling Retrieval Using AKNN}
For each record 
%DC $\recc_{i=1..N_{train}}$ 
$r\in C$ ($C$ is the category dataset), we use AKNN to retrieve the top-$K$ similar siblings from $C$, while making sure that neither of them share the same product with $r$. %Given $N_c$ records in $C$, 
We collect 
%up to $K \cdot (N_c-1)/2$ 
training example pairs 
by coupling each record $r$ with each of its siblings:
$D'(C) = \bigcup_{r_i\in C}\{(r_i, r_j)|r_j \in S(r_i)\}$.

%\paragraph{Approximate K-Nearest Neighbors}
For retrieval we use Universal Sentence Encoder (USE) \cite{universal_sentence_encoder} to embed each question $q_i$ into a 512-length vector $e_i$. We use the Annoy\footnote{https://github.com/spotify/annoy} python library for the implementation of %DC an approximate nearest neighbor algorithm for
efficient AKNN retrieval. In all experiments, 
%we use 50 trees and 
for each record we retrieve the top-K $(K=500)$ similar records, based on the cosine-similarity between the embedding vectors.

\paragraph{Twin Detection Using the Q2Q Model}
For each sibling pair $(r_i,r_j)\in D'(C)$, we use our \qq{} model to score their question-similarity and keep only those with 
%$S_{\qq}(q_i, q_j)>\gamma$ 
$\qq(q_i, q_j)>\gamma$
to yield a collection of twin pairs, $D(C)$. We use $\gamma=0.9$ to ensure only highly similar question pairs.

For our \qq{} model, we apply a standard pre-trained RoBERTa \cite{roberta} classifier. Specifically, we use Hugging-Face base-uncased pre-trained model\footnote{https://github.com/huggingface/transformers} and fine-tune\footnote{We use batch size 32, maximum sequence length of 128, learning rate 5e-5, and 3 epochs.} it for the classification task on our \qq{} dataset\footnote{We only used the examples with full agreement.},
%(See Section~\ref{sec:pqsim}). 
while splitting the data into train, dev and test sets with 80\%-10\%-10\% partition, respectively. For $\gamma=0.5$ (its minimal value) the model achieves test accuracy of 83.2\% with a precision of 81.3\% and a recall of 87.7\%. When setting the twin confidence level threshold to $\gamma=0.9$, the precision of the \qq{} model raises to 89.9\% with a recall of 69.5\%.
%\david{I do not follow - what does it mean setting the confidence level? And how it relates to twin selection through $\gamma$?}
%\avihai{We consider twins only questions where Q2Q(qi,qj) > $\gamma$. Setting higher $\gamma$ value yields greater precision but lower recall of twins.}

We compare the performance of the \qq{} similarity classifier with several unsupervised baselines, namely: (a) Jaccard similarity, (b) cosine similarity over USE embedding, and (c) cosine similarity over RoBERTa\footnote{Hugging-Face sentence-transformers roberta-large-nli-stsb-mean-tokens model.} embedding.
The results are summarized in Table~\ref{tab:q2q_results}, showing that the \qq{} model significantly outperforms these baselines.

\begin{table}[!t]
    \centering
    \small
    \begin{tabular}{ccccc}
    \toprule
    \textbf{Majority} &  \textbf{Jaccard} & \textbf{USE}  &  \textbf{RoBERTa} & \textbf{\qq{}} \\
    \textbf{vote} &  \textbf{similarity} & \textbf{cosine}  &  \textbf{cosine} & \textbf{} \\
    53.1  & 62.0  & 69.6  & 70.7  & 83.2   \\
    \bottomrule
\end{tabular}

    \caption{\capsize{Classification accuracy of  question similarity models.}}
    \label{tab:q2q_results}
\end{table}

\subsection{\cps{} Model }
\label{sec:cps_evaluation}
\paragraph{Training} The \cps{} model predicts the contextual similarity between a pair of twin records. In our experiments, the textual content of a product consists of the product title concatenated with the product bullet points, separated by semicolons. 
%Since a many products do not have any description, we ignored this information. 
The question text is the original query as appeared in the Amazon PQA-dataset.
For the encoding modules of the \cps{} model we use a standard pre-trained RoBERTa-based model as well, while using the $[SEP]$ token for separating the two inputs to each encoder.
For training, twin pairs are labeled according to their contextual similarity using Equation~\ref{eq:labeling}.

We train, fine-tune, and test, an independent CPS model for each category set $C$, using $D(C)$, $D_{dev}(C)$, and $D_{test}(C)$ (details of the data split described in Appendix~\ref{sec:appendix_a}). The training set $D(C)$ is created as described in Section \ref{sec:data_preparation}.
$D_{dev}(C)$ and $D_{test}(C)$, are created the same with one modification -- rather than retrieving the siblings for a record from the dataset it belongs to, the siblings are retrieved from $D(C)$, for both $D_{dev}(C)$, and $D_{test}(C)$. This represents a real-world scenario where existing products with their related questions are used as a corpus for predicting the answer to a question about a new product. Each product with all related questions appear only in one of these sets.
%\avihai{The statistics of the datasets are shown in Table~\ref{tab:prep_dataset}}.
%\david{I guess you mean the sizes of $D(C)$, $D_{dev}(C)$ and $D_{test}(C)$ for the different categories?}

\paragraph{Evaluation}
We evaluate the \cps{} model by measuring the accuracy of its  contextual similarity prediction over $D_{test}(C)$.
The accuracy per category is presented in Table~\ref{tab:cps_results}. 
\begin{table}[!t]
    \centering
    \footnotesize
    
\begin{tabular}{p{2.6cm}ccc}
    \toprule
    \textbf{Category} & \textbf{Acc} &  \textbf{Majority} & \textbf{$\Delta$} \\
    \midrule
%    \rowcolor{lightgray} \multicolumn{5}{c}{Criteria} \\ 
%    \midrule
%    \textbf{Depth* } & 1 - 3 & 4 - 6 & $> 7$ \\
%    \textbf{Length} & 6 - 15 & 16 - 25 & $>26$ \\
    %\rowcolor{lightgray} \multicolumn{4}{c}{placeholder 1} \\ 
    Light Bars         & 73.9 & 61.1 & +12.8 \\
    Monitors           & 78.2 & 68.2 & +9.9  \\
    Smartwatches       & 80.0 &	65.6 & +14.4 \\
    Receivers          & 77.5 & 67.6 & +9.9  \\
    Backpacks          & 83.9 & 76.0 & +7.9  \\
    Jeans              & 71.3 & 59.3 & +11.9 \\
    Beds               & 84.6 & 72.0 & +12.6 \\
    Home Office Desks  & 73.9 & 63.4 & +10.5 \\
    Masks              & 75.1 & 66.9 & +8.2  \\
    Posters \& Prints  & 72.3 & 60.9 & +11.5 \\
    Accessories        & 79.1 & 72.4 & +6.6  \\
    \midrule
    Macro Average   & 77.2 & 66.7 & +9.7  \\    
    \bottomrule

\end{tabular}

    \caption{CPS model test set results on the CPS auxiliary datasets and the majority baseline of each category.}
    \label{tab:cps_results}
\end{table}
The model achieves a relatively high accuracy with a macro average of 77.2\% over all categories, presenting a significant lift of 9.7\% over the majority decision baseline. This is an encouraging result, considering the fact that the answers for many questions cannot be directly inferred from the product textual information. We conjecture that the model is able to learn the affinity between different products, in the context of a given question, for predicting their contextual similarity. 
%An example of two products which are similar in the context of one question but different in context of another one, is provided in Table~\ref{tab:cps_examples}.
For example, the two backpacks {\em Ranvoo Laptop Backpack} and {\em Swiss Gear Bungee Backpack}, were correctly classified by the \cps{} model as similar ($\psi\geq 0.5$) in context of the question ``{\em Will this fit under a plane seat?}'', and classified as different ($\psi<0.5$) in context of the question ``{\em Does it have a separate laptop sleeve?}''.

\subsection{Answer Prediction Methods}
\label{sec:baselines}
We experiment with our \paqa{}
%\footnote{We fine-tuned $w_{min}$ on our dev set for all categories combined as set it to 0.38} 
model and with a few baselines over the test set of all categories. The first one is {\em Majority} which returns the majority answer among all records in the category. Other methods are 
described  next.
%two ablation variants of \paqa{}: 1) Question Similarity Only ignores the twins' contextual similarity with the target record; 2) Product Similarity Only considers only products' textual similarity with the target product. Additionally, we experiment with a standard \apclong{}, as well as with a system that integrates the latter into \paqa{}.
%\ohad{should we also state we check the majority? should we elaborate about what majority is (i.e. always predicting yes?}\avihai{yes, we should. Also explain why the majority drops as number of twins grows.}

\paragraph{\paqa{}}
Given a target record $r_t$,  \paqa{} scores each of its twins by the \cps{} model and predicts the answer for $q_t$, using Equation~\ref{eq:mixture_eq}. $w_{min}$ was fine-tuned on the combined dev set of all categories and was set to 0.38.

\paragraph{\qsolong{} (\qso)}
We modify the \paqa{} model to ignore the CPS classification score when implementing the Mixture-of-Experts model (Eq. \ref{eq:mixture_eq}), by setting an equal weight of $1.0$ to all twin votes:
%\begin{align}
$
    Pred(a_t|r_t)=\textrm{sign}\left(\sum_{r_j\in T(r_t)}\delta(a_j)\right).
$
%\end{align}

\paragraph{\psolong{} (\pso)}
We modify the \paqa{} model by setting $q_t$ and $q_j$ to empty strings at the input of the CPS model, both during training and during inference, forcing it to rely on the products' textual content alone. The twin retrieval process remains untouched. 

\paragraph{\apclong{} (\apc)}
%We compare \paqa{} results with a direct QA prediction approach. In this approach, 
We experiment with a direct prediction approach that only considers the product textual content and the question for answer prediction. For each category $C$, we fine-tune a pre-trained RoBERTa-based classifier over all records $r_j\in C$, using $q_j$ and $p_j$ (separated by the $[SEP]$ token) as input and $\delta(a_j)$ as the training label. 
%For each category we fine-tune and test the baseline separately.\avihai{Full training details are available on Appendix~\ref{sec:appendix}.}

\paragraph{\hybrid{} }
The experimental results show that different answer-prediction methods (e.g. \paqa{} vs \apc{}) may be preferable for different product categories. Therefore, we combine both methods, for achieving optimal results, by mixing the vote of \apc{} with the twin votes, using the Mixture-of-Experts approach:
\[
\begin{array}{ll} 
Pred(a_t|r_t) = \\ \ \ \ \ \ \textrm{sign} \left( \eta(r_t)\delta(\alpha_t) + \sum_{r_j \in T(r_t)}\lambda(r_j)\delta(a_j)\right)
\end{array}
\]
% \[
% \begin{array}{ll}
%     P(a_t|r_t) = \frac{1}{2} + \\ \displaystyle{\frac{1}{2|T(r_t)|+1}\left(\eta(r_t)\delta(\alpha_t) + \sum_{r_j \in T(r_t)}\lambda(r_j)\delta(a_j)\right)}
%     \label{}
% \end{array}
% \]
where $\alpha_t$ is the APC predicted answer, and $\eta(r_t)=\eta_1$, $\eta_2$ and $\eta_3$ for  $|T(r_t)| \leq 10$, $10 < |T(r_t)| < 50$ and $|T(r_t)| \geq 50$, respectively\footnote{We also tried a few different splits on the development set}.
% \[
% \begin{equation} 
% \small
%     \eta(r_t)= 
% \begin{cases}
%     \eta_1,&  |T(r_t)| \leq 10\\
%     \eta_2,&  10 < |T(r_t)| < 50\\
%     \eta_3,&  |T(r_t)| \geq 50.
% \end{cases} \label{eq:eta}
% \end{equation} 
% \]
All $\eta$ values ($\eta>0$) are fine-tuned on the development set for each category separately. The values we used are detailed in Table~\ref{tab:eta} in Appendix~\ref{sec:appendix_a}.

%At inference time, the \apc{} vote is weighted according to the number of twins of $r_t$.

\begin{table}[!t]
    \centering
    \scriptsize
    \begin{tabular}{lcc}
    \toprule
     & \textbf{\# Twins}  &  \textbf{Answer} \\
    %(\jeans) Will these shrink after a wash?  & 100  & 92\% No \\
    %(\jeans) Are these genuine Levis?  & 42  & 90\% Yes \\
    (\monitors) Does this require WiFi? & 51 & 91\% No \\
    (\backpacks) Will it fit under a plane seat? & 213 & 90\% Yes \\
    (\smartwatches) Can it measure blood sugar level?  & 34  & 97\% No \\
    (\smartwatches) Does it come with a charger?  & 269  & 96\% Yes \\
    \bottomrule
    \end{tabular}
    \caption{\capsize{Examples for popular questions with high answer agreement over different products.}}
    \label{tab:similar_questions_examples}
\end{table}

%%%%%%%%%%%%%%%%%%%%%%%%%%%%
\begin{comment}
\subsection{\cps{} Model Evaluation}
\label{sec:cps_evaluation}
The CPS model predicts the contextual similarity of each record in the test set to each record in the corpus. The accuracy results per category are presented in Table~\ref{tab:cps_results}. The model achieves a relatively high accuracy with a macro average of 77.2\% over all eleven categories, presenting a significant lift of 9.7\% over the majority baseline. This is an encouraging result, considering the fact that the answers for many questions cannot be directly extracted from the product textual information. We conjecture that the model is able to exploit relevant features from the text of the products in the context of the given question, and use it as a strong signal for the classification. An example of two products which are similar in context of one questions but different in context of another question, along with the correct prediction of the \cps{} model is provided in Table~\ref{tab:cps_examples}.
\end{comment}

\begin{figure}
   \centering
   \includegraphics[width=3in]{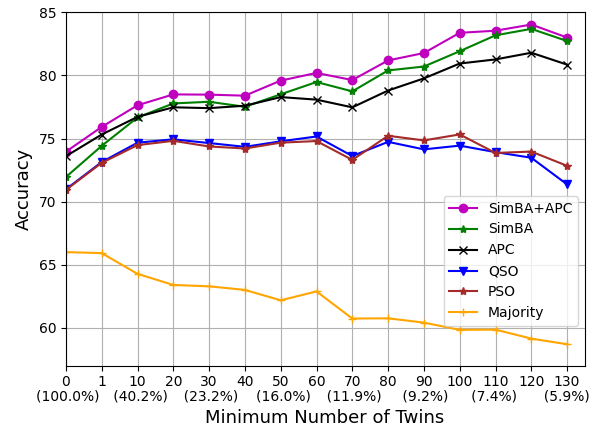}
   \caption{\capsize{Macro-average test accuracy over all categories. The horizontal axis indicate the minimal number of twins and the percentages of test questions each subset represents.}}
   \label{fig:accuracy_vs_nb_twins}
\end{figure}

\begin{table*}[ht]
    \centering
    \scriptsize

\begin{tabular}{c|cccc|cccc}
\toprule
 & \multicolumn{4}{c}{Questions with 1+ Twins}  & \multicolumn{4}{c}{Questions with 60+ Twins} \\ 
 &	\% of data & \paqa{} &	\apc &		\hybrid &	\% of data & \paqa{} &	\apc &	\hybrid \\
\hline
\lightbars  	& 62.5 &  \textbf{75.3}  &	74.8 			&		77.7 &	9.0 	&	\textbf{74.1} 	&	68.1 			&	75.0\\
\monitors   	& 79.2 &  			76.0 &	\textbf{76.4}   &		76.5 &	27.0 	&	78.5        	&	78.5 			&	78.5\\
\smartwatches   & 88.3 &  			77.3 &	\textbf{79.3}   &		79.0 &	31.9 	&			79.9 	&	\textbf{81.7} 	&	80.8 \\
\receivers      & 57.5 &  			70.1 &	\textbf{70.3}   &		72.0 &	4.8 	&	\textbf{83.2} 	&	77.9 			&	83.2\\
\backpacks      & 74.7 &  			80.7 &	\textbf{82.7}   &		82.3 &	21.5 	&	        82.7 	&	\textbf{83.2}	&	82.9\\
\jeans      	& 63.3 &  \textbf{67.4}  &	65.9    		&		67.4 &	13.4 	&	\textbf{74.8} 	&	70.9 			&	75.7\\
\beds   		& 70.4 &  \textbf{77.7}  &	76.4    		&		77.9 &	16.8 	&	\textbf{82.5} 	&	80.4 			&	82.5\\
\home  			& 65.0 &  			71.8 &	\textbf{76.2} 	&		75.8 &	4.7 	&	 		80.0 	&	\textbf{84.4} 	&	84.4 \\
\masks          & 76.5 &  			70.9 &	\textbf{74.2} 	&		73.0 &	4.2 	&	 		80.4 	&	\textbf{82.6} 	&	80.4 \\
\posters  		& 60.1 &  			73.4 &	73.4 			&		74.4 &	9.3 	&	\textbf{75.9} 	&	69.6 			&	75.9\\
\accessories 	& 71.7 &  			78.1 &	\textbf{79.0} 	&		79.2 &	7.2 	&	\textbf{82.3} 	&	81.6 			&	82.7\\
\hline                                                                                                                                   
Macro Average   & 69.9 &  			74.4 &	\textbf{75.3} 	&		75.9 &	13.6 	& \textbf{79.5} 	&	78.1 			&	80.2\\
\bottomrule
\end{tabular}

    \caption{\capsize{Answer prediction accuracy by category. Left: accuracy over records with at least one twin, representing 69.9\% of the records on average. Right: accuracy over records with at least 60 twins, representing 13.6\% of the records. The highest accuracy between \paqa{} and \apc{} is in bold. }}
    \label{tab:results_by_category}
\end{table*}

\subsection{Answer Prediction Evaluation}
\label{sec:paqa_results}
The answer prediction accuracy results of all tested predictors, macro-averaged over $D_{test}(C)$ of all categories, are presented in Figure~\ref{fig:accuracy_vs_nb_twins}. 
%We compare the accuracy of \paqa{} with \apc{} and with the \hybrid{} method that combines the two approaches. We also present results of the two ablation variants \qso{} and \pso{}, and the majority-vote as a reference.
We inspect the performance of the methods on different subsets of the test data, where each subset is determined by all records having at least $x$ twins, $x \in [0..130]$.
The horizontal axis indicates the minimal number of twins in the subset and the percentage of the data each subset represents. 
For example, the results at $x=0$ represent the entire test set, while the results at $x=10$ represents the subset of questions with at least 10 twins, account for 40.2\% of the test set.

The performance of Majority begins with $66\%$ (the percent of `yes' questions in the entire population) and drops for  questions with many twins. We hypothesize  that {\em "obvious"} questions, for which the answer is the same across many products, are rarely asked hence have fewer twins. In contrast, {\em informative} questions, for which the answer is varied across products, are frequently asked w.r.t. many products, hence have many twins. Therefore we see a drop in accuracy of the Majority baseline as the number of twins grows. 

The accuracy of \qso{} is significantly higher than the majority-vote baseline. This demonstrates an interesting phenomena in the  data of 
%certain groups of 
similar questions that tend to have the same answer over variety of products, typically of the same type. A few examples are presented in Table~\ref{tab:similar_questions_examples}. The \qso{} method successfully detects these groups of questions and predicts the majority answer for each such group.
We find that \pso{} method generally doesn't improve over \qso. This is somewhat surprising, as we expected that using product similarity information, such as brand, model, or key features, would increase the prediction accuracy. This demonstrates the importance of question-context, as used in \paqa{}, in addition to the product information alone. 

Moving to \paqa{}, we can see a large performance improvement over the \qso{} and \pso{} methods, which we attribute directly to the \cps{} model. We also see consistent improvement in accuracy with the number of twins, likely due to the larger support the model has for predicting the answer.

The \apc{} method, despite its relative simplicity, performs very well and greatly outperforms the majority-vote and the \qso{} and \pso{} baselines. For the segment of questions with less than 10 twins, \apc{} outperforms the \paqa{} method. This segment represents roughly 60\% of the questions. However, 
%DC: Already been said: as the number of twins grows the improvement of \paqa{} method is increased, and 
for the segment of questions with 60 or more twins, which accounts for 13.6\% of the questions, \paqa{} method consistently outperforms the inductive baseline by 1-2\%. When inspecting the results by category, as shown in Table~\ref{tab:results_by_category}, we can see that considering all questions with at least 1 twin, the \apc{} method dominates in 7 out of the 11 categories, while for questions with at least 60 twins, \paqa{} method dominates in 6 out of the 11 categories.

Finally, we see that the two approaches compliment each other and can be effectively joined, as the \hybrid{} method outperforms both of them over all subsets.

\section{Conclusions}
\label{sec:conclusions}
We presented \paqa{}, a novel answer prediction approach in the PQA domain, which directly leverages similar questions answered with respect to other products.
Our empirical evaluation shows that on some segments of questions, namely those with roughly ten or more similar questions in the corpus, our method can outperform a strong inductive method that directly utilizes the question and the textual product content. We further show that the two approaches are complementary and can be integrated to increase the overall answer prediction accuracy.

For future work, we plan to explore how \paqa{} can be extended and be applied beyond yes-no questions, e.g., for questions with numerical answers or open-ended questions.
Another interesting research direction is combining additional voters to the Mixture-of-Experts model, such as a review-aware answer predictor or a product details-based predictor. Additionally, our current evaluation considered a static view of the answered product-question corpus, we plan to explore temporal aspects of our method, for example, considering questions age or ignoring answers of obsolete products that might be irrelevant.

% in addressing questions for totally new types of products or aspects.
% Ohad: Additionally, since we used already-answered questions for testing our method, we plan to test our method on unanswered questions, considering as twins only other questions asked before the sampling time of the original question.
% ** Avihai's version: Additionally, our current evaluation considered a static view of the answered-Q&As corpus, we plan to explore the effects of temporal aspects on our method, for example, in addressing questions for totally new types of products or aspects.

\section*{Acknowledgments}
Ohad Rozen would like to express his gratitude to Yanai Elazar, Vered Shwartz, and Ido Dagan for providing him valuable advice while he was conducting this research during his internship at Amazon.

\bibliography{anthology,custom}

\begin{thebibliography}{21}
\expandafter\ifx\csname natexlab\endcsname\relax\def\natexlab#1{#1}\fi

\bibitem[{Carmel et~al.(2018)Carmel, Lewin-Eytan, and Maarek}]{Carmel2018}
David Carmel, Liane Lewin-Eytan, and Yoelle Maarek. 2018.
\newblock Product question answering using customer generated content -
  research challenges.
\newblock SIGIR '18, page 1349–1350. Association for Computing Machinery.

\bibitem[{Cer et~al.(2018)Cer, Yang, Kong, Hua, Limtiaco, John, Constant,
  Guajardo{-}Cespedes, Yuan, Tar, Sung, Strope, and
  Kurzweil}]{universal_sentence_encoder}
Daniel Cer, Yinfei Yang, Sheng{-}yi Kong, Nan Hua, Nicole Limtiaco, Rhomni~St.
  John, Noah Constant, Mario Guajardo{-}Cespedes, Steve Yuan, Chris Tar,
  Yun{-}Hsuan Sung, Brian Strope, and Ray Kurzweil. 2018.
\newblock \href {http://arxiv.org/abs/1803.11175} {Universal sentence encoder}.
\newblock \emph{CoRR}, abs/1803.11175.

\bibitem[{Chen et~al.(2019)Chen, Guan, Zhao, Zhao, Wang, Zhao, and
  Sun}]{chen2019answer}
Long Chen, Ziyu Guan, Wei Zhao, Wanqing Zhao, Xiaopeng Wang, Zhou Zhao, and
  Huan Sun. 2019.
\newblock Answer identification from product reviews for user questions by
  multi-task attentive networks.
\newblock In \emph{Proceedings of the AAAI Conference on Artificial
  Intelligence}, volume~33, pages 45--52.

\bibitem[{Cui et~al.(2017)Cui, Huang, Wei, Tan, Duan, and
  Zhou}]{cui2017superagent}
Lei Cui, Shaohan Huang, Furu Wei, Chuanqi Tan, Chaoqun Duan, and Ming Zhou.
  2017.
\newblock Superagent: A customer service chatbot for e-commerce websites.
\newblock In \emph{Proceedings of ACL 2017, System Demonstrations}, pages
  97--102.

\bibitem[{Das et~al.(2019)Das, Wang, Jaffe, Chattopadhyay, Fosler{-}Lussier,
  and Ramnath}]{das2019learning}
Manirupa Das, Zhen Wang, Evan Jaffe, Madhuja Chattopadhyay, Eric
  Fosler{-}Lussier, and Rajiv Ramnath. 2019.
\newblock \href {http://arxiv.org/abs/1910.08270} {Learning to answer
  subjective, specific product-related queries using customer reviews by
  adversarial domain adaptation}.
\newblock \emph{CoRR}, abs/1910.08270.

\bibitem[{Deng et~al.(2020)Deng, Zhang, and Lam}]{Deng2020}
Yang Deng, Wenxuan Zhang, and Wai Lam. 2020.
\newblock \href {https://doi.org/10.1145/3340531.3411904} {Opinion-aware answer
  generation for review-driven question answering in e-commerce}.
\newblock In \emph{Proceedings of CIKM 2020}, page 255–264. Association for
  Computing Machinery.

\bibitem[{Fan et~al.(2019)Fan, Feng, Sun, Li, and Wang}]{fan2019reading}
Miao Fan, Chao Feng, Mingming Sun, Ping Li, and Haifeng Wang. 2019.
\newblock Reading customer reviews to answer product-related questions.
\newblock In \emph{Proceedings of the 2019 SIAM International Conference on
  Data Mining}, pages 567--575. SIAM.

\bibitem[{Gao et~al.(2019)Gao, Ren, Zhao, Zhao, Yin, and
  Yan}]{DBLP:conf/wsdm/GaoRZZYY19}
Shen Gao, Zhaochun Ren, Yihong~Eric Zhao, Dongyan Zhao, Dawei Yin, and Rui Yan.
  2019.
\newblock \href {https://doi.org/10.1145/3289600.3290992} {Product-aware answer
  generation in e-commerce question-answering}.
\newblock In \emph{Proceedings WSDM 2019}, pages 429--437. {ACM}.

\bibitem[{He and Dai(2011)}]{google-he}
Jing He and Decheng Dai. 2011.
\newblock \href {http://proceedings.mlr.press/v20/he11.html} {Summarization of
  yes/no questions using a feature function model}.
\newblock volume~20 of \emph{Proceedings of Machine Learning Research}, pages
  351--366, South Garden Hotels and Resorts, Taoyuan, Taiwain. PMLR.

\bibitem[{Jacobs et~al.(1991)Jacobs, Jordan, Nowlan, and
  Hinton}]{mixture_of_experts}
Robert~A. Jacobs, Michael~I. Jordan, Steven~J. Nowlan, and Geoffrey~E. Hinton.
  1991.
\newblock \href
  {http://dblp.uni-trier.de/db/journals/neco/neco3.html#JacobsJNH91} {Adaptive
  mixtures of local experts.}
\newblock \emph{Neural Comput.}, 3(1):79--87.

\bibitem[{Jeon et~al.(2005)Jeon, Croft, and Lee}]{jeon2005finding}
Jiwoon Jeon, W~Bruce Croft, and Joon~Ho Lee. 2005.
\newblock Finding similar questions in large question and answer archives.
\newblock In \emph{Proceedings CIKM 2005}, pages 84--90.

\bibitem[{{Lai} et~al.(2018){Lai}, {Bui}, {Lipka}, and {Li}}]{Lai2018}
T.~M. {Lai}, T.~{Bui}, N.~{Lipka}, and S.~{Li}. 2018.
\newblock \href {https://doi.org/10.1109/ICMLA.2018.00180} {Supervised transfer
  learning for product information question answering}.
\newblock In \emph{Proceedings of ICMLA 2018}, pages 1109--1114.

\bibitem[{Liu et~al.(2019)Liu, Ott, Goyal, Du, Joshi, Chen, Levy, Lewis,
  Zettlemoyer, and Stoyanov}]{roberta}
Yinhan Liu, Myle Ott, Naman Goyal, Jingfei Du, Mandar Joshi, Danqi Chen, Omer
  Levy, Mike Lewis, Luke Zettlemoyer, and Veselin Stoyanov. 2019.
\newblock \href {http://arxiv.org/abs/1907.11692} {Roberta: {A} robustly
  optimized {BERT} pretraining approach}.
\newblock \emph{CoRR}, abs/1907.11692.

\bibitem[{McAuley(2016)}]{amazon-qa-data}
Julian McAuley. 2016.
\newblock Amazon question/answer data.
\newblock \url{https://jmcauley.ucsd.edu/data/amazon/qa/"}.

\bibitem[{McAuley and Yang(2016)}]{DBLP:conf/www/McAuleyY16}
Julian~J. McAuley and Alex Yang. 2016.
\newblock \href {https://doi.org/10.1145/2872427.2883044} {Addressing complex
  and subjective product-related queries with customer reviews}.
\newblock In \emph{Proceedings WWW 2016}, pages 625--635. {ACM}.

\bibitem[{Park et~al.(2015)Park, Kim, Zhai, and
  Guo}]{DBLP:conf/sigir/ParkKZG15}
Dae~Hoon Park, Hyun~Duk Kim, ChengXiang Zhai, and Lifan Guo. 2015.
\newblock \href {https://doi.org/10.1145/2766462.2767748} {Retrieval of
  relevant opinion sentences for new products}.
\newblock In \emph{Proceedings SIGIR 2015}, pages 393--402. {ACM}.

\bibitem[{Pourgholamali(2016)}]{Pourgholamali16}
Fatemeh Pourgholamali. 2016.
\newblock \href {https://doi.org/10.1145/2959100.2959102} {Mining information
  for the cold-item problem}.
\newblock In \emph{Proceedings of RecSys 2016}, page 451–454. ACM.

\bibitem[{Wan and McAuley(2016)}]{wan2016modeling}
Mengting Wan and Julian McAuley. 2016.
\newblock Modeling ambiguity, subjectivity, and diverging viewpoints in opinion
  question answering systems.
\newblock In \emph{Proceeding of ICDM 2016}, pages 489--498. IEEE.

\bibitem[{Yu et~al.(2018)Yu, Qiu, Jiang, Huang, Song, Chu, and Chen}]{Yu2018}
Jianfei Yu, Minghui Qiu, Jing Jiang, Jun Huang, Shuangyong Song, Wei Chu, and
  Haiqing Chen. 2018.
\newblock \href {https://doi.org/10.1145/3159652.3159685} {Modelling domain
  relationships for transfer learning on retrieval-based question answering
  systems in e-commerce}.
\newblock In \emph{Proceedings of WSDM 2018}, page 682–690. ACM.

\bibitem[{Yu et~al.(2012)Yu, Zha, and Chua}]{yu2012answering}
Jianxing Yu, Zheng-Jun Zha, and Tat-Seng Chua. 2012.
\newblock Answering opinion questions on products by exploiting hierarchical
  organization of consumer reviews.
\newblock In \emph{Proceedings of EMNLP 2012}, pages 391--401. ACL.

\bibitem[{Yu and Lam(2018)}]{DBLP:conf/wsdm/YuL18}
Qian Yu and Wai Lam. 2018.
\newblock \href {https://doi.org/10.1145/3159652.3159718} {Review-aware answer
  prediction for product-related questions incorporating aspects}.
\newblock In \emph{Proceedings of WSDM 2018}, pages 691--699. {ACM}.

\end{thebibliography}
\bibliographystyle{acl_natbib}

\clearpage
\appendix
\section{Supplemental Material}
\label{sec:appendix_a}
\subsection{\amazonpqsim{} dataset}
\label{sec:pqsim_dataset}
The Amazon-PQSim dataset includes question pairs, where all questions are published on Amazon website. Each pairs has a corresponding label: 1 for similar, else 0. The labels were collected via Appen crowd sourcing service. 
We took the following filtering steps (step b in ~\ref{sec:pqsim}) for each question: 
\begin{itemize}
\item  Removed any question with less than five words.
\item Removed any question with more than 15 words. 
\item  Removed any none-English questions. 
\item Removed any question with multiple question-marks (may indicate multiple questions).
\item Removed questions with rare words (any word which is not in the top 2000 most frequent words). 
\end{itemize}

\subsection{\amazonpqa{} dataset}
\label{sec:pqa_dataset}
The \amazonpqa{} dataset includes questions and their answers that are published on Amazon website, along with the public product information and category (Amazon \textit{Browse Node name}). The data includes the following fields:
\begin{itemize}
%\item `answer\_id`,  
\item `question\_id`,  
\item `asin\_id`,  
\item `question\_text`,  
\item `answer\_text`,  
%\item `answer\_helpful\_votes`,  
%\item `answer\_unhelpful\_votes`, 
%\item `question\_helpful\_votes`, 
%\item `question\_unhelpful\_votes`,  
%\item `badges` (is-seller, is-vendor), 
%\item `question\_submit\_date`,  
%\item `answer\_submit\_date`, 
%\item `question\_language\_code`, 
%\item `answer\_language\_code`,  
\item `bullet\_points`,  
\item `product\_description`,  
\item `brand\_name`,  
\item `item\_name `, 
%\item `browse\_node\_path`. 
\end{itemize}
 In addition, we augment this data with fields derived from our current work: 
 \begin{itemize}
\item `is\_yes-no\_question`,  
\item `yes-no\_answer` (yes, no, maybe),
\end{itemize}

\paragraph{Yes/No  Question Classification}
\label{sec:appendix_a_yes_no_question_classification}
We followed  \cite{google-he} for detecting Yes/No questions using simple heuristics, such as checking if the question starts with a \textit{Be verb} (am, is, are, been, being, was, were), \textit{Modal verb} (can, could, shall, should, will, would, may, might) or an \textit{Auxiliary verb} (do, did, does, have, had, has), and additionally ends with a question mark. We tested the classifier on  McAuley's dataset~\cite{amazon-qa-data}, 
identified yes/no questions with 98.4\% precision at 96.5\% recall, while considering McAuley as ground truth\footnote{\citeauthor{DBLP:conf/www/McAuleyY16} reported identifying yes/no questions with 97\% precision at 82\% recall on their dataset.}.

\paragraph{Yes/No  Answer  Labeling}
\label{sec:appendix_a_yes_no_answer_labeling}
As described in Section~\ref{sec:pqa_dataset}, we used the McAuley dataset~\cite{amazon-qa-data} to train a RoBERTa-based classifier, taking only yes/no questions. For testing, we used 5 annotator to annotate 583 question-answer pairs, randomly sampled from our raw data. The model achieved 97\% and 88\% precision for the \textit{yes} and \textit{no} labels, respectively, and a recall of 65\% and 51\% on the entire test set\footnote{\citeauthor{DBLP:conf/www/McAuleyY16} reported 98\% accuracy after keeping only the 50\% of instances about which their classifier
was the most confident.}.

Next, to determine each question's final yes/no answer, we first omitted answers classified as \textit{maybe}. When a question is answered by a verified seller, we considered it as most reliable and used it as the final label. Otherwise we used the majority votes among the remaining answers. In our experiments, we ignore questions with an equal number of \textit{yes} and \textit{no} answers.

\paragraph{Dataset Split}
Each item in our dataset is a (product, question, answer) triplet. We split the labeled triplets into train (80\%), dev (10\%), and test (10\%) sets for each category, relating to the number of products. Each product with all related questions appear only in one of these sets. The statistics for this dataset are given in Table~\ref{tab:pqa_stats}.
% (For more details see Table~\ref{tab:} in Appendix~\ref{sec:appendix_a}).

\subsection{CPS Model Details}
The CPS has a total of 254.6M parameters. For all incorporated RoBERTa models we use a maximum sequence length of 256, dropout of 0.1 , and a 32 batch size for training. We applied different learning rates and number of epochs for each product-category. The specific values we used after tuning are shown in Table~\ref{tab:cps_hyperparams}.

\begin{table*}[ht]
    \centering
    \scriptsize
    
\begin{tabular}{p{2.6cm}cccccc}
    \toprule[1.5pt]
    \textbf{Category} & \textbf{\# Products} &  \textbf{\# Questions} & \textbf{\# Y/N Questions} & \textbf{\# Answers} & \textbf{\# Q. Answered Yes} & \textbf{\# Q. Answered No} \\
    \midrule
    Light Bars         &  6,151     & 48,736     &  23,956     &  95,853    &  10,146    &  5,243   \\
    Monitors           &  6,246     & 170,529    &  86,837     &  316,126   &  35,353    &  22,947  \\
    Smartwatches       &  8,611     & 166,400    &  94,936     &  289,945   &  41,683    &  22,033  \\
    Receivers          &  3,442     & 58,618     &  33,511     &  135,700   &  14,488    &  7,364   \\
    Backpacks          &  12,096    & 68,598     &  38,914     &  138,996   &  19,902    &  6,090   \\
    Jeans              &  38,008    & 61,908     &  17,518     &  129,346   &  7,708     &  5,548   \\
    Beds               &  17,202    & 108,723    &  46,722     &  238,786   &  17,757    &  13,917  \\
    Home Office Desks  &  6,986     & 55,303     &  23,202     &  112,958   &  9,523     &  5,971   \\
    Masks              &  13,384    & 51,295     &  24,989     &  100,789   &  9,757     &  5,759   \\
    Posters \& Prints  &  33,803    & 53,939     &  20,737     &  99,926    &  8,171     &  5,450   \\
    Accessories        &  38,825    & 238,603    &  159,272    &  438,447   &  60,990    &  23,772  \\
    \midrule                                                                                   
    Rest of 89 Categories & 1,288,754 	& 8,906,362 &   4,833,639  & 18,565,933 &  2,219,022 & 1,055,816 \\
    \midrule[1pt]
    Total              & 1,473,508 	& 9,989,014 &   5,404,233  & 20,662,805 &  2,454,500 & 1,179,910 \\
	\bottomrule[1.5pt]
\end{tabular}

% \begin{tabular}{p{2.6cm}ccccc}
%     \toprule
%     \textbf{Category} & \textbf{\# Products} &  \textbf{\# Questions} & \textbf{\# Answers} & \textbf{\# Questions per Product} & \textbf{\# Answers per Question} \\
%     \midrule
%     Light Bars         & 6,187    & 49,220    & 97,010    & 8    & 2   \\
%     Monitors           & 6,364    & 172,167   & 319,717   & 27.1 & 1.9 \\
%     Smartwatches       & 8,637    & 166,843   & 290,897   & 19.3 & 1.7 \\
%     Receivers          & 3,481    & 59,126    & 137,164   & 17   & 2.3 \\
%     Backpacks          & 12,263   & 69,389    & 140,835   & 5.7  & 2   \\
%     Jeans              & 39,426   & 64,112    & 134,928   & 1.6  & 2.1 \\
%     Beds               & 17,401   & 110,398   & 243,247   & 6.3  & 2.2 \\
%     Home Office Desks  & 7,060    & 56,350    & 115,749   & 8    & 2.1 \\
%     Masks              & 13,466   & 51,578    & 101,405   & 3.8  & 2   \\
%     Posters \& Prints  & 34,229   & 54,576    & 101,291   & 1.6  & 1.9 \\
%     Accessories        & 39,399   & 241,485   & 445,107   & 6.1  & 1.8 \\
%     \midrule
%     Total            & 187,913  & 1,095,244 & 2,127,350 &	5.8* & 1.9* \\    
%     \bottomrule

% \end{tabular}

    \caption{\capsize{Statistics of the \amazonpqa{} dataset extracted from Amazon.com. \textit{\# Y/N Questions} as detected by our Yes/No Question detector;  \textit{\# Answers} is the total number of answers before any filtering; \textit{\# Q. Answers Yes} (\textit{No}) is the number of Yes/No questions with answers labeled as \textit{Yes} (\textit{No})  } (See Section~\ref{sec:pqa})}
    \label{tab:extracted_data_stats}
\end{table*}

\begin{table*}
    \centering
    \scriptsize
    
\begin{tabular}{p{2cm}|ccc|ccc|ccc|ccc}                                                                 
    \toprule                                                                                   
    & \multicolumn{3}{c}{Train Set}  & \multicolumn{3}{c}{Dev Set} & \multicolumn{3}{c}{Test Set} & \multicolumn{3}{c}{Total} \\ 
    \textbf{Category} & \textbf{\# P} &  \textbf{\# Q} & \textbf{\% Yes} & \textbf{\# P} &  \textbf{\# Q} & \textbf{\% Yes} & \textbf{\# P} &  \textbf{\# Q} & \textbf{\% Yes} & \textbf{\# P} &  \textbf{\# Q} & \textbf{\% Yes} \\
    \midrule                                                                                   
            Light bars	&   2,552  &    8,675	 & 68.1		 &   319	    &    1,080 & 	68.7 &        319	   &     1,296	 & 69.2	 & 	      3,190	 &      11,051	 & 68.3	\\
            Monitors	&   3,421  &   29,886	 & 63.3		 &   427	    &    3,890 & 	64.7 &        427	   &     4,260	 & 63.0	 & 	      4,275	 &      38,036	 & 63.4	\\
        Smartwatches	&   4,128  &   34,734	 & 68.5		 &   516	    &    3,730 & 	66.4 &        516	   &     3,778	 & 67.8	 & 	      5,160	 &      42,242	 & 68.3	\\
           Receivers	&   1,725  &   11,991	 & 69.2		 &   215	    &    1,827 & 	68.1 &        215	   &     2,356	 & 65.7	 & 	      2,155	 &      16,174	 & 68.5	\\
           Backpacks	&   4,834  &   14,740	 & 78.4		 &   604	    &    1,397 & 	75.9 &        604	   &     1,908	 & 77.3	 & 	      6,042	 &      18,045	 & 78.0	\\
               Jeans	&   5,365  &    6,056	 & 61.3		 &   670	    &      773 & 	59.8 &        670	   &      769	 & 58.1	 & 	      6,705	 &       7,598	 & 60.8	\\
                Beds	&   5,912  &   16,792	 & 59.1		 &   739	    &    2,017 & 	58.3 &        739	   &     2,276	 & 58.3	 & 	      7,390	 &      21,085	 & 58.9	\\
   Home Office Desks	&   2,576  &    8,637	 & 62.7		 &   322	    &    1,059 & 	64.3 &        322	   &      962	 & 62.9	 & 	      3,220	 &      10,658	 & 62.9	\\
               Masks	&   4,332  &    8,541	 & 64.8		 &   541	    &    1,180 & 	64.0 &        541	   &     1,099	 & 63.1	 & 	      5,414	 &      10,820	 & 64.5	\\
   Posters \& Prints	&   5,739  &    7,226	 & 62.7		 &   717	    &     ,868 & 	62.4 &        717	   &      850	 & 66.0	 & 	      7,173	 &       8,944	 & 63.0	\\
         Accessories	&  14,422  &   54,125	 & 73.5		 &  1,802	    &    6,038 & 	73.7 &       1,802	   &     6,706	 & 74.5	 & 	     18,026	 &      66,869	 & 73.6	\\
    \midrule                                                                     
    Total  		        & 55,006   & 201,403 & 66.5* & 6,872 & 23,859 & 66.0* & 6,872 & 26,260 & 66.0* &	 68,750 & 251,522 & 66.4* \\                       
    \bottomrule                                                                                
                                                                                               
\end{tabular}

    \caption{\capsize{Statistics of the yes-no questions subset from the \amazonpqa{} dataset, and the train, dev, test splits used in our experiments. Only categories used for our experiments are displayed. *Macro average}}
    \label{tab:pqa_stats}
\end{table*}

%\begin{table}[!t]
\begin{table}[ht]
    \centering
    \scriptsize
    
\begin{tabular}{p{2.6cm}cc}
    \toprule
    \textbf{Category} & \textbf{\# Epochs} &  \textbf{Learning Rate} \\
    \midrule

   Light Bars         &         3 &        3.0E-5 \\
   Monitors           &         4 &        3.0E-5 \\
   Smartwatches       &         3 &        3.0E-5 \\
   Receivers          &         4 &        3.0E-5 \\
   Backpacks          &         4 &        3.0E-5 \\
   Jeans              &         3 &        2.0E-5 \\
   Beds               &         4 &        4.0E-5 \\
   Home Office Desks  &         3 &        2.0E-5 \\
   Masks              &         3 &        3.0E-5 \\
   Posters \& Prints  &         3 &        2.0E-5 \\
   Accessories        &         3 &        2.0E-5 \\

    \bottomrule

\end{tabular}

    \caption{\capsize{Number of epochs and learning rates used for training the CPS model (Section~\ref{sec:cps_evaluation}) on each category}}
    \label{tab:cps_hyperparams}
\end{table}

%\begin{table}[!t]
\begin{table}[ht]
    \centering
    \footnotesize
    
\begin{tabular}{p{2.6cm}ccc}
    \toprule
    \textbf{Category} & \textbf{$\eta_1$} &  \textbf{$\eta_2$} & \textbf{$\eta_3$} \\
    \midrule

   Light Bars         &          3 &          8 &          2   \\
   Monitors           &          6 &          2 &         63   \\
   Smartwatches       &          2 &         11 &         49   \\
   Receivers          &          2 &          0 &          0   \\
   Backpacks          &          1 &          4 &         17   \\
   Jeans              &          7 &          8 &         22   \\
   Beds               &          1 &          0 &          0   \\
   Home Office Desks  &          4 &          1 &         38   \\
   Masks              &          4 &          6 &          3   \\
   Posters \& Prints  &          5 &          0 &         18   \\
   Accessories        &          1 &          2 &         16   \\

    \bottomrule

\end{tabular}

    \caption{$\eta_1$, $\eta_2$ and $\eta_3$ values used after fine-tuning on our development set (Section~\ref{sec:baselines}). Larger $\eta$ values give more weight to the \apc{} vote.}
    \label{tab:eta}
\end{table}

\end{document}